\documentclass{article}
\usepackage{ijcai24}

\usepackage{times}
\usepackage{soul}
\usepackage{url}
\usepackage[hidelinks]{hyperref}
\usepackage[utf8]{inputenc}
\usepackage[small]{caption}
\usepackage{graphicx}
\usepackage{amsmath}
\usepackage{amsthm}
\usepackage{booktabs}
\usepackage{algorithm}
\usepackage{algorithmic}
\usepackage[switch]{lineno}

\usepackage[T1]{fontenc}
\usepackage{graphicx}
\usepackage{algorithm}
\usepackage{algorithmic, mathrsfs, bm}
\usepackage{multirow}
\usepackage{array}
\usepackage{amsmath}
\usepackage{appendix}
\usepackage{float}
\usepackage{footmisc}
\usepackage{colortbl}
\usepackage{cleveref}

\usepackage{subcaption}
\usepackage{footnote}

\usepackage{tablefootnote}
\usepackage{scrextend}

\usepackage{hyperref}

\usepackage{multirow}

\usepackage{threeparttable}

\usepackage{amssymb}

\newtheorem{lemma}{Lemma}

\title{Effective Gradient Sample Size via Variation Estimation for Accelerating Sharpness aware Minimization}

\author{
Jiaxin Deng$^1$
\and
Junbiao Pang$^1$\and
Baochang Zhang$^2$\and
Tian Wang$^2$ \\
\affiliations
$^1$Beijing University of Technology\\
$^2$Beihang University\\
\emails
dengjiaxin@emails.bjut.edu.cn,
junbiao\_pang@bjut.edu.cn,
bczhang@@buaa.edu.cn,
wangtian@buaa.edu.cn
}

\preto{\abstractkeywords}{\nolinenumbers}

\begin{document}

\maketitle

\begin{abstract}
Sharpness-aware Minimization (SAM) has been proposed recently to improve model generalization ability.
However, SAM calculates the gradient twice in each optimization step, thereby doubling the computation costs compared to stochastic gradient descent (SGD).
In this paper, we propose a simple yet efficient sampling method to significantly accelerate SAM.
Concretely, we discover that the gradient of SAM is a combination of the gradient of SGD and the Projection of the Second-order gradient matrix onto the First-order gradient (PSF).
PSF exhibits a gradually increasing frequency of change during the training process.
To leverage this observation, we propose an adaptive sampling method based on the variation of PSF, and we reuse the sampled PSF for non-sampling iterations.
Extensive empirical results illustrate that the proposed method achieved state-of-the-art accuracies comparable to SAM on diverse network architectures.
\end{abstract}

\section{Introduction}\label{sec:intro}

The powerful generalization ability of Deep Neural Networks (DNNs) has led to significant success in many fields~\cite{chaudhari-2019-Entropy_sgd-IOP,izmailov-2018-SWA-UAI,kwon-2021-asam-ICML}.
In recent years, some research has been proposed to understand the generalization of DNNs~\cite{keskar-2016-large_batch-ICLR,zhang-2021-understanding-ACM,mulayoff-2020-unique_properties-ICML,andriushchenko-2022-towards_understanding-ICML,zhou-2021-towards_understanding-NIPS,zhou-2022-understanding_the_robustness-ICML}.
Several studies have verified the relationship between flat minima and generalization error~\cite{dinh-2017-sharp_minima-ICML,li-2018-visualizing-NIPS,jiang-2019-fantastic-ICLR,liu-2020-loss-NIPS,sun-2021-exploring-AAAI}.
Among these studies, Jiang et al.~\cite{jiang-2019-fantastic-ICLR} explored over 40 complexity measures and demonstrated that a sharpness-based measure exhibits the highest correlation with generalization.
\begin{figure}
  \centering
  \includegraphics[width=0.68\linewidth]{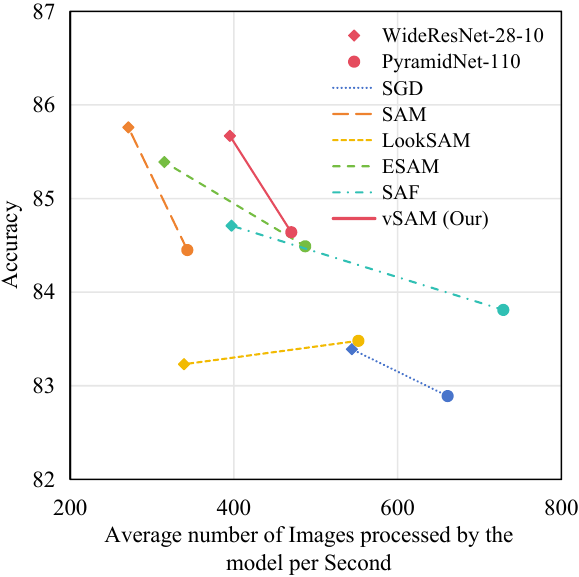}
  \caption{Accuracy vs training speed of SGD, SAM, LookSAM, ESAM, SAF and vSAM (Our). Every connected line represents a method that trains WideResNet-28-10 and PyramidNet-110 models on CIFAR-100.
  vSAM substantially accelerates training with almost no reduction in accuracy.
  }
  \label{fig:result}
\end{figure}

Based on the connection between sharpness of the loss landscape and model generalization, Foret et al. proposes Sharpness Aware Minimization (SAM)~\cite{foret-2020-SAM-ICLR} to seek out parameter values whose entire neighborhoods have uniformly low training loss value.
SAM minimizes both the loss value and the loss sharpness to obtain a flat minimum and improve model generalization.
SAM and its variants have demonstrated state-of-the-art performance across various applications~\cite{kwon-2021-asam-ICML,du-2022-ESAM-ICLR,liu-2022-looksam-CVPR,chen-2022-vision_transformer-ICLR,zheng-2021-regularizing-CVPR,zhuang-2022-GSAM-ICLR}.
However, SAM requires two forward and backward operations in one iteration, which results in SAM's optimization speed being only half that of SGD.
In some scenarios, dedicating twice the training time to achieve only a marginal improvement in accuracy may not strike an optimal balance between accuracy and efficiency.
For example, when SAM is employed to optimize WideResNet-28-10 on CIFAR-100, despite achieving a higher test accuracy than SGD (84.45\% vs. 82.89\%), the optimization speed is only half that of SGD (343 imgs/s vs. 661 imgs/s), as illustrated in Figure~\ref{fig:result}.

\begin{figure*}
  \centering
  \begin{subfigure}{0.32\linewidth}
		\centering
		\includegraphics[width=0.9\linewidth]{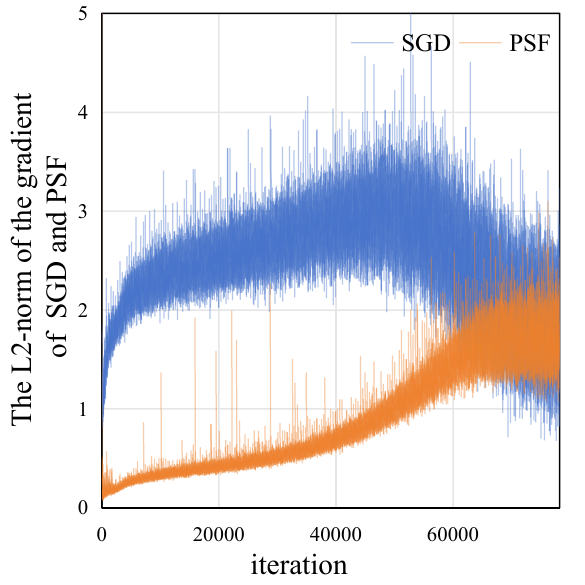}
		\caption{Resnet-18}
		\label{fig:fig1a}
  \end{subfigure}
  \centering
  \begin{subfigure}{0.32\linewidth}
		\centering
		\includegraphics[width=0.9\linewidth]{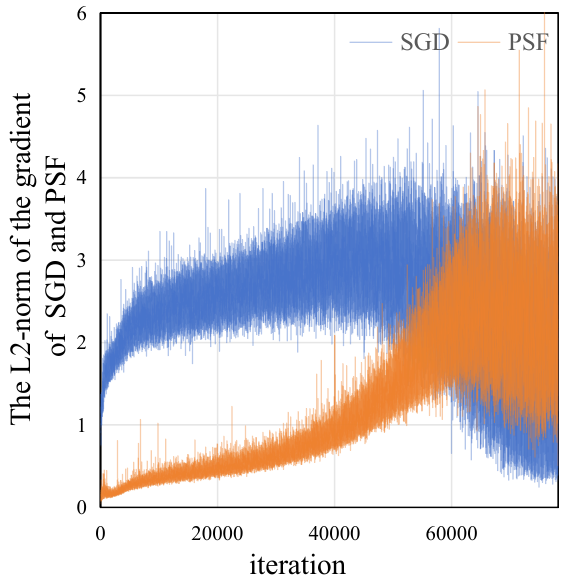}
		\caption{WideResNet-28-10}
		\label{fig:fig1b}
  \end{subfigure}
  \centering
  \begin{subfigure}{0.32\linewidth}
		\centering
		\includegraphics[width=0.9\linewidth]{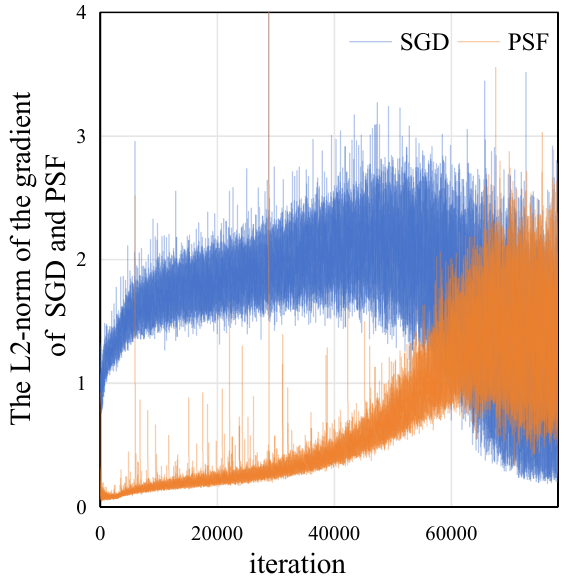}
		\caption{PyramidNet-110}
		\label{fig:fig1c}
  \end{subfigure}
  \caption{The variation trend of $||\nabla L_i^{SGD}||$ and $||\nabla L_i^{PSF}||$ during training. $||\nabla L_i^{SGD}||$ and $||\nabla L_i^{PSF}||$ are denotes the L2-norm of the gradient of SGD and the L2-norm of PSF, respectively. The trends of $||\nabla L_i^{SGD}||$ and $||\nabla L_i^{PSF}||$ are similar across different models (Resnet-18, WideResNet-28-10, PyramidNet-110).}
  \label{fig:l2norm}
\end{figure*}

In this paper, we accelerate SAM by introducing an adaptive gradient sampling strategy, ensuring nearly no drop in accuracy.
Specifically, we have observed that the gradient of SAM can be decomposed into two parts: the first part is the gradient of SGD, which searches for a local minimum, while the second part is the Projection of the Second-order gradient matrix onto the First-order gradient (PSF), which drives SAM to search for a flat region.
As illustrated in Fig.~\ref{fig:l2norm}, the L2-norm of PSF (hereinafter referred to as L2-PSF) gradually increases with iterations, and its amplitude changes from small values to large ones.
This indicates that the efficacy of the PSF varies throughout the training process.
Motivated by the observation in Fig.~\ref{fig:l2norm}, we propose to accelerate SAM by adaptively sampling the PSF.
Concretely, during the sampling iterations, we compute the PSF, while during the non-sampling iterations, we reuse the previously computed PSF.
By controlling the sampling rate, we can significantly speed up the optimization process by reducing the computation of the PSF.


In this paper, we propose Variation-based SAM (vSAM) to adaptively adjust the sampling rate based on the variance of the L2-PSF. 
The variance is a natural way to measure the sampling ratio, reflecting the difference of the PSF between two iterations.
We increase the sampling rate of the PSF to ensure that the model's generalization ability does not degrade when the PSF changes significantly, while we reduce the sampling rate of the PSF when the gradient of SGD dominates the optimization.
In this way, we speed up the optimization while maintaining the generalization performance of the model.

In a nutshell, our contributions are as follows:
\begin{itemize}
    \item To the best of our knowledge, we firstly find that the gradient of SAM can be decomposed into the gradient of SGD and PSF. 
          Furthermore, we observe a gradual increase in the L2-norm of PSF across various networks during training, with its amplitude changing from small to large values.
          This observation motivates us to adaptively sample the PSF to accelerate SAM.
    \item We introduce the vSAM, which adaptively samples and reuses gradients to not only preserve the model's generalization ability but also accelerate the optimization efficiency. 
          The empirical results demonstrate that vSAM achieves a 40\% acceleration compared to SAM, while ensuring almost no decrease in model generalization ability.    
\end{itemize}

\section{Related Works}
\label{sec:related}
\textbf{Background of SAM.}
The notion of seeking minima that are "flat minima" can be credited to~\cite{hochreiter1994simplifying}, and there has been considerable research examining its relationship with generalization~\cite{zhang-2021-understanding-ACM,keskar-2016-large_batch-ICLR,wei-2019-improved-ICLR,jiang-2019-fantastic-ICLR}.
Based on these studies, Foret et al.~\cite{foret-2020-SAM-ICLR} introduced an effective approach to improving the model's generalization ability, called SAM.
The optimization process of SAM can be viewed as addressing a minimax optimization problem, formulated as follows:
\begin{equation}\label{equ:sam}
\begin{aligned}
\mathop {\min } \limits_\mathbf{w} & \; {L^{SAM}}(\mathbf{w}) + \lambda ||\mathbf{w}||_2^2 \\
& where \; {L^{SAM}}(\mathbf{w}) = \mathop {\max }\limits_{||\bm{\varepsilon} || \le \rho } L(\mathbf{w} + \bm{\varepsilon} ),
\end{aligned}
\end{equation}
where $\bm{\varepsilon}$ represents weight perturbations in Euclidean ball with radius $\rho$, $L^{SAM}$ is the perturbed loss, $\lambda ||\mathbf{w}||_2^2$ is a standard L2 regularization term.
In order to minimize $L^{SAM}(\mathbf{w})$, SAM utilizes Taylor expansion to search for the maximum perturbed loss in local parameter space:
\begin{equation}\label{equ:psf}
\begin{aligned}
\mathop {\arg \max }\limits_{||\bm{\varepsilon} || \le \rho } \; L(\mathbf{w} + \bm{\varepsilon} ) & \approx \mathop {\arg \max }\limits_{||\bm{\varepsilon} || \le \rho } \; L(\mathbf{w}) + {\bm{\varepsilon} ^T}{\nabla _\mathbf{w}}L(\mathbf{w}) \\
&= \mathop {\arg \max }\limits_{||\bm{\varepsilon} || \le \rho } \; {\bm{\varepsilon} ^T}{\nabla _\mathbf{w}}L(\mathbf{w}),
\end{aligned}
\end{equation}
By solving Eq.~\eqref{equ:psf}, SAM can calculate the gradient once to obtain the perturbation $\hat{ \bm{\varepsilon} } = \rho {\nabla _\mathbf{w}}L(\mathbf{w})/||{\nabla _\mathbf{w}}L(\mathbf{w})||$ that can maximize the loss function.
Minimizing the loss of the perturbed weight $\mathbf{w} + \hat{ \bm{\varepsilon} }$ promotes the entire neighborhood of the weight $\mathbf{w}$ to have low training loss values. 
Through gradient approximation, the optimization problem of SAM is reduced to:
\begin{equation}\label{equ:grad_approx}
\begin{aligned}
\mathop {\min }\limits_\mathbf{w} \; {L^{SAM}}(\mathbf{w}) \approx \mathop {\min }\limits_\mathbf{w} \; L(\mathbf{w} + \hat{ \bm{\varepsilon} }),
\end{aligned}
\end{equation}
Finally, SAM requires a second calculation of the gradient to optimize the model, as follows:
\begin{equation}\label{equ:grad_sam}
\begin{aligned}
{\nabla _\mathbf{w}}{L^{SAM}}(\mathbf{w}) \approx {\nabla _\mathbf{w}}L(\mathbf{w} + \bm{\hat\varepsilon} ) \approx {\nabla _\mathbf{w}}L(\mathbf{w}){|_{\mathbf{w} + \bm{\hat\varepsilon} }},
\end{aligned}
\end{equation}
From Eq.~\eqref{equ:psf} and Eq.~\eqref{equ:grad_sam}, we observe that SAM necessitates two forward and backward operations to update weights once.

\noindent\textbf{Efficient optimization methods for SAM.}
Some methods have been proposed to enhance the optimization speed of SAM, broadly categorized into two groups.

The first category of methods focuses on reducing the data and model weights involved in training.
For example, Du et al. propose ESAM~\cite{du-2022-ESAM-ICLR}, which employs two training strategies, Stochastic Weight Perturbation (SWP) and Sharpness Sensitive Data Selection (SDS), to reduce computation.
SWP approximates the weight perturbation by using a subset of model weights, while SDS updates the model using a subset of training data that contributes the most to sharpness.
However, SWP barely accelerate training speed since that gradients should be computed for almost every weight, while SDS still requires a forward and backward in each optimization step.

The second category of methods focuses on reducing the computation of gradients. 
For example, Du et al. propose SAF and MESA~\cite{du-2022-saf-NIPS}, which employ trajectory loss to approximate sharpness through a surrogate sharpness measure (the loss of SGD between several iterations).
SAF and MESA lacks direct weight perturbations to assist SAM in escaping local minimum regions (as discussed in Experiments section).
Additionally, SAF requires a large amount of memory to store the output results for all data, while MESA also needs to maintain an Exponential Moving Average (EMA) model during training.
Liu et al. propose LookSAM~\cite{liu-2022-looksam-CVPR} which periodically calculates the gradient of SAM every 5 iterations.
This periodic strategy ensures that the majority of optimization steps do not require two forward and backward passes.
However, this uniform gradient sampling fails to reflect the importance of sharpness optimization in the overall optimization process.
Subsequently, Jiang et al. claims that the SAM update is more useful in sharp regions than in flat regions.
As a result, they design AE-SAM~\cite{jiang-2022-aesam-ICLR} to adaptively employ SAM based on the loss landscape geometry.
However, it requires sharpness to be evaluated at each optimization step.
In this paper, we propose an adaptive sampling strategy to reflect the importance of sharpness and adjust the sampling rate of the PSF.

\section{Method}
\subsection{Gradient Composition of SAM}
We rewrite Eq.~\eqref{equ:grad_sam} based on Taylor expansion and substitute $\bm{\hat \varepsilon}$ to obtain the following:
\begin{equation}\label{equ:method_taylor_expansion}
\begin{aligned}
{\nabla _\mathbf{w}}L(\mathbf{w} + \bm{\hat\varepsilon} ) &\approx {\nabla _\mathbf{w}}(L(\mathbf{w}) + \bm{\hat \varepsilon} {\nabla _\mathbf{w}}L(\mathbf{w}))\\
 &= {\nabla _\mathbf{w}}(L(\mathbf{w}) + \rho ||{\nabla _\mathbf{w}}L(\mathbf{w}){\rm{||)}}.
\end{aligned}
\end{equation}
The gradient of SAM can be considered as a combination of SGD's gradient and the gradient of the L2-norm of SGD's gradient.
We expand the second term in Eq.~\eqref{equ:method_taylor_expansion} as follows:
\begin{equation}\label{equ:method_exp_grad}
{\nabla _\mathbf{w}}(\rho ||{\nabla _\mathbf{w}}L(\mathbf{w}){\rm{||)}} = \rho \frac{{{\nabla _\mathbf{w}}L(\mathbf{w}) \nabla _\mathbf{w}^2L(\mathbf{w})}}{{||{\nabla _\mathbf{w}}L(\mathbf{w}){\rm{||}}}}.
\end{equation}
The L2-norm of SGD's gradient (hereinafter referred to as L2-SGD) can be transformed into the Projection of the Second-order gradient matrix onto the First-order gradient (PSF).
Eq.~\eqref{equ:method_taylor_expansion} and Eq.~\eqref{equ:method_exp_grad} indicate that the gradient of SAM can be decomposed into the gradient of SGD and PSF.
The PSF prompts SAM to find a flat minimum.

\begin{lemma} \label{lemma1}
Let $\nabla _\mathbf{w}^2L(\mathbf{w})$ be a positive definite matrix with n eigenvalues, then the L2-PSF has an upper bound as follows:
\begin{equation}\label{equ:method_lemma}
\left\| {{\nabla _\mathbf{w}}(\rho \left\| {{\nabla _\mathbf{w}}L(\mathbf{w})} \right\|)} \right\| \le \rho \sum\limits_{i = 1}^n {{\delta _i} |\cos (\theta_i )|}, 
\end{equation}
\end{lemma}
\noindent where $\delta _i$ is the $i$-th eigenvalue,  $\theta_i$ is the angle between $U_i$ and $\nabla _\mathbf{w} L(\mathbf{w})$, $U_i$ is the $i$-th eigenvector of $\nabla _\mathbf{w}^2L(\mathbf{w})$.

We observe that the L2-PSF gradually increases during training, and its amplitude changes from small values to large ones, as shown in Figure~\ref{fig:l2norm}.
From Lemma~\ref{lemma1}, we can see that the L2-PSF is related to the eigenvalues of matrix $\nabla _\mathbf{w}^2L(\mathbf{w})$, as well as the angle formed between the eigenvectors and the gradient $\nabla _\mathbf{w} L(\mathbf{w})$.
The changes in the L2-PSF may be attributed to more pronounced variations in the direction of $\nabla _\mathbf{w} L(\mathbf{w})$ during the later stages of optimization.
The observed phenomenon reflects the evolving significance of the PSF during the training process, prompting us to utilize the variance of the L2-PSF (i.e.,$||\nabla L_i^{PSF}||=||{\nabla L_i^{SAM}-\nabla L_i^{SGD}}||$) for effective control of gradient sampling during training.
We adjust the sampling rate according to the following rules:
\begin{itemize}
\item When the PSF changes slowly, the PSF in the current iteration can be replaced by the PSF in the previous iteration. 
We decrease the sampling rate to alleviate the computational burden of the PSF.
\item When the PSF changes rapidly, we enhance the sampling rate to calculate the PSF in a high frequency.
This ensures that the gradient is similar to the original SAM, thereby improving the generalization ability of the model.
\end{itemize}

\subsection{Effective Gradient Sample Size via Variation Estimation}
\textbf{Adaptive gradient sampling based on variance.}
In order to estimate the variance of $||\nabla L^{PSF}||$, we save $||\nabla L^{PSF}||$ in the previous $N$ sampled iterations and denote as $\mathbf{g}^{norm}$.
However, there might be individual values in $\mathbf{g}^{norm}$ that are extremely large or small, as shown in Figure~\ref{fig:l2norm}.
Using the overall variance of these L2-norm values directly could impact the variation of the sampling rate.
Therefore, we propose sort $\mathbf{g}^{norm}$ in ascending order and denote the result as $\mathbf{\hat{g}}^{norm}$.
The sorted values are then evenly divided into $M$ slices and denoted as $[\mathbf{\hat{g}}^{norm}_1,...,\mathbf{\hat{g}}^{norm}_M]$.
Finally, we calculate the variance of L2-norm values within each slice and compute the average of these variances as follows:
\begin{equation}\label{equ:method_v}
\begin{aligned}
v_i = \frac{1}{M}\sum\limits_{j = 1}^{M} var(\mathbf{\hat{g}}^{norm}_j),
\end{aligned}
\end{equation}
where $var(\mathbf{g})$ is a function for calculating the variance of $\mathbf{g}$, $v_i$ evaluates the variance of the PSF in the $i$-th iteration.
The change rate of the variance at $i$-th iteration can be written as follows:
\begin{equation}\label{equ:method_c_var_i}
\begin{aligned}
c^{var}_i = \frac{v_i-v_{i-1}}{v_{i-1}}.
\end{aligned}
\end{equation}
To more accurately evaluate changes in variance of the previous $N$ sampled iterations, we calculate the change rate according to Eq.~\eqref{equ:method_c_var_i} at each sampled iteration and average it as follows:
\begin{equation}\label{equ:method_c_var}
\begin{aligned}
\hat{c}^{var} = \frac{1}{{N-1}}\sum\limits_{i = 2}^{N} {{c^{var}_i}}.
\end{aligned}
\end{equation}

\noindent \textbf{Adaptive gradient sampling based on norm value.}
When the gradient of SGD dominates the training, we aim to speed up the optimization by reducing the computation of the PSF.
We use the ratio of $||\nabla L_{i}^{PSF}||$ to $||\nabla L_{i}^{SGD}||$ at the $i$-th iteration as a factor controlling the sampling rate, as follows:
\begin{equation}\label{equ:method_r}
\begin{aligned}
r_{i} = \frac{||\nabla L_{i}^{PSF}||}{||\nabla L_{i}^{SGD}||}.
\end{aligned}
\end{equation}
In Eq.~\eqref{equ:method_r}, $r_{i}$ reflects the relative importance between PSF and SGD during the optimization process.
The change rate of the ratio at the $i$-th iteration and average change rate in the previous $N$ sampled iterations can be written as Eq.~\eqref{equ:method_c_norm_i} and Eq.~\eqref{equ:method_c_norm}, respectively.
\begin{align}
c^{norm}_i &= \frac{r_i-r_{i-1}}{r_{i-1}}, \label{equ:method_c_norm_i}\\
\hat{c}^{norm} &= \frac{1}{{N-1}}\sum\limits_{i = 2}^{N} {{c^{norm}_i}}. \label{equ:method_c_norm}
\end{align}

We adjust the sampling rate in an autoregressive manner, the sampling rate for the next $N$ iterations be written as follows:
\begin{equation}\label{equ:method_s}
s_{i+1}= s_{i} \cdot (1+ \alpha \cdot\hat{c}^{var}+ \alpha \cdot \hat{c}^{norm}),
\end{equation}
\begin{equation}\label{equ:method_p}
p_{i+1} = \frac{s_{i+1}}{N},
\end{equation}
where $s_{i}$ is the number of sampling required in the last $N$ iterations, $\alpha$ is a hyperparameter that can adjust the variation of the sampling rate, and $p_{i+1}$ is the sampling rate in the next $N$ iterations. 
In the next $N$ iterations, the PSF is sampled with a probability of $p_{i+1}$ in each iteration.
In Eq.~\eqref{equ:method_s}, we can see that the number of samples required for the next $N$ iterations will be controlled by $\hat{c}^{var}$ and $\hat{c}^{norm}$.
In addition, the computational cost of L2-norm becomes larger as the model size increases.
We observe that the L2-SGD and the L2-PSF in the last few layers of the model also capture the changes in the gradient of SGD and PSF.
Thus, we replace $||\nabla L^{SGD}||$ and $||\nabla L^{PSF}||$ with the L2-SGD and the L2-PSF in the last few layers of the model, respectively. 

\begin{algorithm}[tb]
    \caption{{Pseudocode of the proposed vSAM}}
    {\bf Require:}
    The training dataset, the learning rate $\eta$, parameters $\alpha$, $\gamma$, $\rho$, $N$, $M$.
    \begin{algorithmic}[1]
    \STATE Initialization: $s_1$, iteration $I_{start}$ to start applying the adaptive sampling rate.
    \FOR{$i = 1,2,\cdot\cdot\cdot$}
    \STATE Calculating the gradient of SGD $\nabla L^{SGD}_i$;
    \IF{$i <= I_{start}$ \OR sampling}
    \STATE Calculating the gradient of SAM $\nabla L^{SAM}_i$;
    \STATE Calculating the PSF $\nabla L^{PSF}_i$;
    \STATE ${\mathbf{w}_{i+1}} = {\mathbf{w}_{i}} - \eta \nabla L_i^{SAM}$;
    \ELSE
    \STATE ${\mathbf{w}_{i+1}} = {\mathbf{w}_{i}} - \eta (\nabla L_i^{SGD} + {{\gamma} ^{i-i^*}} \cdot \nabla L_{i^*}^{PSF})$;
    \ENDIF
    \IF{$i > I_{start}$ \AND $i$ \% $N == 0$}
    \STATE compute sampling rate $p$;
    \ENDIF
    \ENDFOR
    \end{algorithmic}
    \label{algorithm:1}
\end{algorithm}

\noindent \textbf{Gradient reuse in the non-sampling iteration.} 
In the sampling iteration, the optimization rule can be written as follows:
\begin{equation}\label{equ:sampling_opt}
\centering
\begin{aligned}
{\mathbf{w}_{i+1}} = {\mathbf{w}_{i}} - \eta \nabla L_i^{SAM}
\end{aligned}
\end{equation}
where $\eta$ is learning rate. Eq.~\eqref{equ:sampling_opt} is essentially the optimization rule for SAM.
In the non-sampling iteration, we reuse the PSF from the last sampling iteration, and the optimization rule can be written as follows:
\begin{equation}\label{equ:non-sampling_opt}
\centering
\begin{aligned}
{\mathbf{w}_{i+1}} = {\mathbf{w}_{i}} - \eta (\nabla L_i^{SGD} + {{\gamma} ^{i-i^*}} \cdot \nabla L_{i^*}^{PSF})
\end{aligned}
\end{equation}
where $i^*$ represents the last sampling in the $i^*$-th iteration, $\nabla L_{i^*}^{PSF}$ is the PSF at the last sampling iteration, and $\gamma$ is a hyperparameter that reduces the magnitude of the PSF. 
The reliability of the PSF obtained from the last sampling diminishes progressively as the current iteration distances itself from the last sampled iteration.
Algorithm \ref{algorithm:1} shows the overall proposed algorithm.

\noindent \textbf{Convergence analysis.}
We further analyze the convergence properties of vSAM as follows.
\begin{lemma} \label{lemma2}
Suppose $L^{SGD}(\mathbf{w})$ is $\tau$-Lipschitz smooth and $|L^{SGD}(\mathbf{w})|$ is bounded by $M$. For any $t \in \{0,...,T\}$ and any $\mathbf{w} \in W$, suppose we can obtain bounded observations as follows:
\begin{equation}\label{equ:bound_obs}
\centering
\begin{aligned}
&\mathbb{E}[\nabla L_t^{SGD}(\mathbf{w})]= \nabla L^{SGD}(\mathbf{w}), \|\nabla L_t^{SGD}(\mathbf{w})\|\leq G_1, \\
&\mathbb{E}[\nabla L_t^{PSF}(\mathbf{w})]= \nabla L^{PSF}(\mathbf{w}), \|\nabla L_t^{PSF}(\mathbf{w})\|\leq G_2.
\end{aligned}
\end{equation}
Then with learning rate $\eta_t = \frac{\eta_0}{t}$, we have the following bound for vSAM:
\begin{equation}\label{equ:convergence}
\centering
\begin{aligned}
\frac{1}{T}\sum\limits_{t=1}^T &{\mathbb{E}[||\nabla {L^{SGD}}(\mathbf{w}) + \gamma\nabla {L^{PSF}}(\mathbf{w})||^2]} \\ 
& \le {C_1} + {C_2}InT
\end{aligned}
\end{equation}
where $C_1$ and $C_2$ are constants that only depend on $\tau$, $\gamma$, $\eta_0$, $M$, $G_1$ and $G_2$.
\end{lemma}
It can be seen from Lemma~\ref{lemma2} that the convergence of vSAM is affected by the parameter $\gamma$, the learning rate $\eta_0$, etc., while they are all controlled within a certain range.

\section{Experimental Results}
\subsection{Setup}
\textbf{Datasets.}
To verify the effectiveness of vSAM, we conduct experiments on CIFAR-10 and CIFAR-100 \cite{krizhevsky2009learning} image classification benchmark datasets. 

\noindent\textbf{Models.}
We employ a variety of architectures, i.e. ResNet-18 \cite{he2016deep}, WideResNet-28-10 \cite{zagoruyko2016wide} and PyramidNet-110 \cite{han2017deep} to evaluate the performance and training efficiency.

\noindent\textbf{Baselines.}
We take the vanilla SGD and SAM \cite{foret-2020-SAM-ICLR} as baselines.
To comprehensively evaluate the performance of vSAM, we have also chosen some efficient methods SAM-5, SAM-10, ESAM \cite{du-2022-ESAM-ICLR}, and LookSAM \cite{liu-2022-looksam-CVPR}, SAF \cite{du-2022-saf-NIPS}, MESA \cite{du-2022-saf-NIPS} and AE-SAM \cite{jiang-2022-aesam-ICLR} for comparison.
SAM-5 and SAM-10 indicate that SAM is used every 5 and 10 iterations respectively, while SGD is used for other iterations.
Other efficient methods are the follow-up works of SAM that aim to enhance efficiency.
We re-implemented the SAM, SAM-5, SAM-10, LookSAM, ESAM and SAF in Pytorch.

\noindent\textbf{Implementation details.}
On CIFAR-10 and CIFAR-100, we train all the models 200 epochs using a batch size of 128 with cutout regularization \cite{devries2017improved} and cosine learning rate decay \cite{loshchilov2016sgdr}.
For the proposed method, we set $\gamma=0.7$ for ResNet-18, $\gamma=0.9$ for WideResNet-28-10 and PyramidNet-110, and use $K=50$, $M=5$ for all experiments.
For other hyperparameters, we follow the setup in ESAM. 
We implement vSAM in Pytorch and train models on a single NVIDIA GeForce RTX 3090 three times, each with different random seed.
The adaptive gradient sampling will be deployed after a predefined iteration $I_{start}$, because the variance and norm value of the PSF are neither stable nor reliable at the first few epochs.

Additionally, to avoid using the PSF in every iteration, we set the maximum sampling rate to $0.8$ and halt sampling when the sampling number reaches $0.8 \times N$ within the current $N$ iterations.

The number of images processed per second may change during training.
Therefore, optimization efficiency is evaluated by calculating costs and quantified by the Average number of Images processed by the model per Second (AIS) as follows:
\begin{equation}\label{equ7}
AIS = \frac{D \cdot E}{T},
\end{equation}
where $D$ represents the amount of data for one epoch of training, $T$ is the total training time in seconds, including both data loading and model optimization time, and $E$ represents the training epoch.

\begin{table}[]
\center
\begin{tabular}{c|ccc}
\toprule
$\alpha$ & Accuracy & Sampling number & AIS \\ \midrule
SAM                    &     84.45    &    78200           &     343(100\%)   \\  \midrule
0.05                   &     83.25    &    5224            & \bf{596(174\%)}  \\
0.1                    &     84.01    &    15287           &     461(134\%)   \\
0.13                   &     84.52    &    26466           &     484(141\%)   \\
0.15                   &     84.28    &    39276           &     406(118\%)   \\
0.2                    & \bf{84.66}   &    35150           &     444(129\%)   \\\bottomrule
\end{tabular}
\caption{Parameter Study of $\alpha$ on CIFAR-100 datasets with WideResNet. The best accuracy and efficiency are bolded.}
\label{table_alpha}
\end{table}

\begin{table}[]
\center
\begin{tabular}{c|ccc}
\toprule
$\gamma$ & Accuracy & Sampling number & AIS \\ \midrule
0.6                 &     83.93        &    21219           & \bf{508}   \\
0.7                 &     84.52        &    26372           &     486    \\
0.8                 &     84.52        &    26466           &     484    \\
0.9                 & \bf{84.64}       &    28288           &     469    \\ \bottomrule
\end{tabular}
\caption{Parameter Study of $\gamma$ on CIFAR-100 datasets with WideResNet. The best accuracy and efficiency are bolded.}
\label{table_gamma}
\end{table}

\begin{table*}
\center
\begin{tabular}{c|ccc|ccc}
\toprule
           & \multicolumn{3}{c|}{\bf{CIFAR-10}}                                                        & \multicolumn{3}{c}{\bf{CIFAR-100}}                                                        \\ \midrule
\bf{ResNet}     & Accuracy $\uparrow$ & \begin{tabular}[c]{@{}c@{}}Sampling\\ number\end{tabular} $\downarrow$ & AIS $\uparrow$            & Accuracy $\uparrow$ & \begin{tabular}[c]{@{}c@{}}Sampling\\ number\end{tabular} $\downarrow$ & AIS $\uparrow$            \\ \midrule
SGD                & 96.13\scriptsize{$\pm$0.03}                & \textbackslash{}   & 3167(178\%) 
                   & 79.22\scriptsize{$\pm$0.12}                & \textbackslash{}   & 3194(174\%) \\ \midrule
SAM                & 96.62\scriptsize{$\pm$0.04}    & 78200              & 1780(100\%)  
                   & \underline{80.66\scriptsize{$\pm$0.06}}    & 78200              & 1836(100\%) \\
SAM-5              & 96.40\scriptsize{$\pm$0.04}                & 15640              & 2780(156\%) 
                   & 80.42\scriptsize{$\pm$0.09}                & 15640              & 2763(151\%) \\
SAM-10             & 96.28\scriptsize{$\pm$0.05}                & 7820               & 2982(168\%) 
                   & 80.27\scriptsize{$\pm$0.06}                & 7820               & 2969(162\%) \\ \midrule
LookSAM-5          & 96.33\scriptsize{$\pm$0.04}                & 15640              & 2106(118\%) 
                   & 80.21\scriptsize{$\pm$0.05}                & 15640              & 2186(119\%) \\
ESAM               & 96.55\scriptsize{$\pm$0.03}                & \textbackslash{}   & 1895(107\%) 
                   & 80.01\scriptsize{$\pm$0.12}                & \textbackslash{}   & 1919(105\%) \\
ESAM\textsuperscript{1} 
                   & 96.56\scriptsize{$\pm$0.08}                & \textbackslash{}   & 2409(140\%) 
                   & 80.41\scriptsize{$\pm$0.10}                & \textbackslash{}   & 2423(140\%) \\
SAF                & 96.26\scriptsize{$\pm$0.02}                & \textbackslash{}   & 2696(152\%) 
                   & 79.93\scriptsize{$\pm$0.04}                & \textbackslash{}   & 2686(146\%) \\
SAF\textsuperscript{2} 
                   & 96.37\scriptsize{$\pm$0.02}                & \textbackslash{}   & 3213(194\%) 
                   & 80.06\scriptsize{$\pm$0.05}                & \textbackslash{}   & 3248(192\%) \\
MESA\textsuperscript{2} 
                   & 96.24\scriptsize{$\pm$0.02}                & \textbackslash{}   & 2780(168\%) 
                   & 79.79\scriptsize{$\pm$0.09}                & \textbackslash{}   & 2793(165\%) \\
AE-SAM\textsuperscript{3} 
                   & \underline{96.63\scriptsize{$\pm$0.04}}                & 39178              & \textbackslash{} 
                   & 80.48\scriptsize{$\pm$0.11}                & 38944              & \textbackslash{} \\
\midrule
\bf{vSAM}          &\bf{96.64\scriptsize{$\pm$0.12}}            & 23579              & 2492(140\%) 
                   &\bf{80.76\scriptsize{$\pm$0.15}}            & 27377              & 2470(135\%) \\ \midrule
\bf{WideResNet} & Accuracy $\uparrow$ & \begin{tabular}[c]{@{}c@{}}Sampling\\ number\end{tabular} $\downarrow$ & AIS $\uparrow$           & Accuracy $\uparrow$ & \begin{tabular}[c]{@{}c@{}}Sampling\\ number\end{tabular} $\downarrow$ & AIS $\uparrow$            \\ \midrule
SGD                & 96.89\scriptsize{$\pm$0.02}                & \textbackslash{}   & 666(197\%)  
                   & 82.89\scriptsize{$\pm$0.06}                & \textbackslash{}   & 661(193\%) \\ \midrule
SAM                & \underline{97.43\scriptsize{$\pm$0.04}}           & 78200              & 338(100\%)  
                   & 84.45\scriptsize{$\pm$0.05}    & 78200              & 343(100\%) \\
SAM-5              & 97.16\scriptsize{$\pm$0.03}                & 15640              & 553(164\%)  
                   & 83.06\scriptsize{$\pm$0.08}                & 15640              & 549(160\%) \\
SAM-10             & 96.94\scriptsize{$\pm$0.02}                & 7820               & 605(179\%)  
                   & 83.07\scriptsize{$\pm$0.08}                & 7820               & 608(177\%) \\ \midrule
LookSAM-5          & 97.05\scriptsize{$\pm$0.03}                & 15640              & 537(159\%)  
                   & 83.48\scriptsize{$\pm$0.06}                & 15640              & 552(161\%) \\
ESAM               & \bf{97.44\scriptsize{$\pm$0.03}}                & \textbackslash{}   & 480(142\%)  
                   & 84.49\scriptsize{$\pm$0.05}                & \textbackslash{}   & 487(142\%) \\
ESAM\textsuperscript{1} 
                   & 97.29\scriptsize{$\pm$0.11}                & \textbackslash{}   & 550(139\%) 
                   & 84.51\scriptsize{$\pm$0.01}                & \textbackslash{}   & 545(139\%) \\
SAF                & 97.08\scriptsize{$\pm$0.04}                & \textbackslash{}   & 628(186\%) 
                   & 82.33\scriptsize{$\pm$0.03}                & \textbackslash{}   & 633(185\%) \\
SAF\textsuperscript{2} 
                   & 97.08\scriptsize{$\pm$0.15}                & \textbackslash{}   & 727(198\%) 
                   & 83.81\scriptsize{$\pm$0.04}                & \textbackslash{}   & 729(197\%) \\
MESA\textsuperscript{2} 
                   & 97.16\scriptsize{$\pm$0.23}                & \textbackslash{}   & 617(168\%) 
                   & 83.59\scriptsize{$\pm$0.24}                & \textbackslash{}   & 625(169\%) \\
AE-SAM\textsuperscript{3} 
                   & 97.30\scriptsize{$\pm$0.10}                & 38709              & \textbackslash{} 
                   & \underline{84.51\scriptsize{$\pm$0.11}}                & 38787              & \textbackslash{} \\ \midrule
\bf{vSAM}          & 97.29\scriptsize{$\pm$0.08}                & 33251              & 468(139\%)  
                   &\bf{84.64\scriptsize{$\pm$0.17}}            & 29356              & 470(137\%)  \\ \midrule
\bf{PyramidNet} & Accuracy $\uparrow$ & \begin{tabular}[c]{@{}c@{}}Sampling\\ number\end{tabular} $\downarrow$ & AIS $\uparrow$           & Accuracy $\uparrow$ & \begin{tabular}[c]{@{}c@{}}Sampling\\ number\end{tabular} $\downarrow$ & AIS $\uparrow$           \\ \midrule
SGD                & 97.04\scriptsize{$\pm$0.03}                & \textbackslash{}   & 544(200\%)  
                   & 83.39\scriptsize{$\pm$0.04}                & \textbackslash{}   & 544(201\%) \\ \midrule
SAM                &97.72\scriptsize{$\pm$0.02}                 & 78200              & 272(100\%)  
                   &\bf{85.76\scriptsize{$\pm$0.06}}            & 78200              & 271(100\%) \\
SAM-5              & 97.57\scriptsize{$\pm$0.05}                & 15640              & 447(164\%)  
                   & 84.25\scriptsize{$\pm$0.05}                & 15640              & 457(169\%) \\
SAM-10             & 97.25\scriptsize{$\pm$0.04}                & 7820               & 492(181\%) 
                   & 84.20\scriptsize{$\pm$0.08}                & 7820               & 495(183\%) \\ \midrule
LookSAM-5          & 97.23\scriptsize{$\pm$0.03}                & 15640              & 358(132\%)  
                   & 83.23\scriptsize{$\pm$0.08}                & 15640              & 339(125\%) \\
ESAM               & 97.59\scriptsize{$\pm$0.04}                & \textbackslash{}   & 311(114\%)  
                   & 85.39\scriptsize{$\pm$0.09}                & \textbackslash{}   & 315(116\%) \\
ESAM\textsuperscript{1} 
                   & \underline{97.81\scriptsize{$\pm$0.01}}    & \textbackslash{}   & 401(139\%) 
                   & 85.56\scriptsize{$\pm$0.05}                & \textbackslash{}   & 381(138\%) \\
SAF                & 97.07\scriptsize{$\pm$0.05}                & \textbackslash{}   & 488(179\%) 
                   & 83.23\scriptsize{$\pm$0.06}                & \textbackslash{}   & 489(180\%) \\
SAF\textsuperscript{2} 
                   & 97.34\scriptsize{$\pm$0.06}                & \textbackslash{}   & 391(202\%) 
                   & 84.71\scriptsize{$\pm$0.01}                & \textbackslash{}   & 397(200\%) \\
MESA\textsuperscript{2} 
                   & 97.46\scriptsize{$\pm$0.09}                & \textbackslash{}   & 332(171\%) 
                   & 84.73\scriptsize{$\pm$0.14}                & \textbackslash{}   & 339(171\%) \\
AE-SAM\textsuperscript{3} 
                   & \bf{97.90\scriptsize{$\pm$0.05}}           & 39256              & \textbackslash{} 
                   & 85.58\scriptsize{$\pm$0.10}                & 38944              & \textbackslash{} \\ \midrule
\bf{vSAM}          & 97.49\scriptsize{$\pm$0.08}                & 29104              & 411(151\%) 
                   & \underline{85.67\scriptsize{$\pm$0.06}}    & 26972              & 395(146\%)  \\ \bottomrule

\multicolumn{7}{l}{\small $1$ We report the results in \cite{du-2022-ESAM-ICLR}. But failed to reproduce them using the officially released codes.}\\                 
\multicolumn{7}{l}{\small $2$ We report the results in \cite{du-2022-saf-NIPS}. For SAF, we failed to reproduce it using the official released code on } \\
\multicolumn{7}{l}{\small CIFAR-10 and CIFAR-100, so we reproduced it ourselves following the algorithmic flow of SAF.} \\                   
\multicolumn{7}{l}{\small $3$ We report the results in \cite{jiang-2022-aesam-ICLR}.}\\                   

\end{tabular}
\caption{The results of the proposed method and the comparison methods on CIFAR-10 and CIFAR-100 dataset.
The numbers in parentheses (·) indicate the ratio of corresponding method's training speed to SAM’s.
$\uparrow$ means that the larger the reported results are better, and $\downarrow$ means that the smaller the results are better.
The best accuracy is in bold and the second best is underlined.}
\label{table_result}
\end{table*}

\subsection{Parameter Studies}
\noindent\textbf{Parameter study of }$\bm{\alpha}$.
We study the effect of $\alpha$ on accuracy and optimization efficiency using the WideResNet-28-10 model on the CIFAR-100 dataset.
The results are presented in Table~\ref{table_alpha}.
As $\alpha$ increases gradually, we observe a corresponding increase in the sampling number, leading to a slowdown in training speed.
However, the accuracy is improved with the sampling number increases.

Moreover, the maximum sampling number does not mean the best accuracy.
It indicates that sampling at the opportune moments is more crucial than sampling number.
For example, vSAM shows a significant improvement in optimization speed compared to SAM and achieves better accuracy than SAM-5 with nearly identical sampling number.
Across all our experiments, we set $\alpha=0.1$ for CIFAR-10 and $\alpha=0.13$ for CIFAR-100 to ensure that vSAM improves the optimization speed by about 40\% compared to SAM

\noindent\textbf{Parameter study of }$\bm{\gamma}$.
We study the effect of $\gamma$ on accuracy and optimization efficiency using the WideResNet-28-10 model on the CIFAR-100 dataset.
The results are presented in Table~\ref{table_gamma}.
We observe that $\gamma$ has a small effect on optimization speed, and gradient reuse plays a limited role when $\gamma$ is small.
Therefore, we recommend using a large value for $\gamma$ in most of cases.

\subsection{Comparison to SOTAs}
We perform experiments to train ResNet-18, WideResNet-28-10, and PyramidNet-110 on CIFAR-10 and CIFAR-100.
The experimental results are presented in Table~\ref{table_result}. 
We evaluate the proposed method and comparative methods from two aspects: classification accuracy and optimization efficiency.

\noindent\textbf{Accuracy.}
From Table~\ref{table_result}, we observe that vSAM achieves significantly higher accuracy compared to SGD, SAM-5, and SAM-10. 
The accuracy of vSAM is comparable to or even surpasses that of SAM.
Furthermore, in comparison to LookSAM, ESAM, and AE-SAM, our results outperform theirs in most cases. 
This demonstrates that vSAM can successfully preserve the model's generalization ability during the training process.
Because vSAM calculates the PSF in more appropriate optimization iterations, it promotes generalization.
vSAM reuses the PSF term in the non-sampling iteration also guarantees the generalization ability of the model.
We also achieve superior results compared to SAF and MESA.
Because SAF and MESA optimize the sharpness from multiple previous iteration to the current iteration, which is different from the sharpness of the current iteration.
SAF would overlook the local sharpness because multiple local sharpness tends to occur the multiple optimization iterations~\cite{zhang-2023-gradient-CVPR}.

\noindent\textbf{Efficiency.}
These experiments demonstrate that vSAM exhibits about 40\% faster training than SAM while achieving comparable accuracy.
While SAM-5 and SAM-10 enhance optimization speed, they exhibit a significant decrease in accuracy.
In certain scenarios, vSAM exhibits a faster optimization speed compared to LookSAM and ESAM.
This is because LookSAM involves additional operations, such as computing the gradient projection, while ESAM still requires computing the gradient twice for one optimization step.
SAF sacrifices memory for optimization efficiency, achieving nearly the same speed as SGD.
MESA accelerates training by utilizing an EMA model and eliminating the need for backward when updating the EMA model.
While vSAM may not match the speed of SAF and MESA, it excels in preserving the model's generalization capability.
Furthermore, vSAM achieves comparable performance to AE-SAM with fewer SAM updates.

\begin{table}[]
\center
\begin{tabular}{ccccc}
\toprule
datasets   & \multicolumn{2}{c}{CIFAR-10}                                   & \multicolumn{2}{c}{CIFAR-100}                          \\ \midrule
methods    & Acc.  & AIS   & Acc.  & AIS \\ \midrule
SAM-5      & 96.40 & 2780(156\%)  & 80.42 & 2763(151\%)                                        \\
vSAM       & 96.64 & 2492(140\%)  & 80.76 & 2470(135\%)                                        \\
vSAM-A     & 96.60 & 2405(135\%)  & 80.60 & 2459(134\%)                                        \\ \bottomrule
\end{tabular}
\caption{Ablation Study of vSAM on CIFAR-10 and CIFAR-100.}
\label{table_ablation}
\end{table}

\subsection{Ablation Study}
To better understand the effectiveness of adaptive sampling rate and gradient reuse in improving the performance and efficiency, we consider a variant of vSAM: 
vSAM with only adaptive sampling rate, optimized with SGD during no-sampling iterations (vSAM-A).
The results are presented in Table~\ref{table_ablation}.

Compared to SAM-5, optimization speed of vSAM-A is slower because the adaptive sampling rate strategy lets the total sampling number of vSAM-A is larger than SAM-5.
However, the accuracy of vSAM-A is higher than that of SAM-5.
This indicates that the adaptive sampling rate can effectively enhance optimization speed while preserving the model's generalization ability.
Comparing vSAM with vSAM-A, we observe that the gradient reuse strategy can also prevent the model's generalization ability without increasing the computational cost.

\subsection{Application to Quantization-Aware Training}
Neural network quantization reduces computational demands by lowering weight and activation precision, enabling efficient deployment on edge devices without compromising model performance~\cite{esser2020learned,wei2021qdrop,nagel2022overcoming}. 

Learned Step size Quantization (LSQ)~\cite{esser2020learned} is a commonly used scheme in quantization tasks.
The quantization parameter step size is set as a learnable parameter to participate in network training instead of calculating by weight distribution.
We employ vSAM as the optimization method for LSQ to demonstrate its broader applicability.
We utilized SGD, SAM, and vSAM to quantize the parameters of ResNet-18 and MobileNets to W4A4 on the CIFAR-10 dataset, and the results are presented in Table~\ref{table_LSQ}.
The training speed of vSAM is over 50\% faster than that of SAM, while maintaining results close to SAM. 
This demonstrates that vSAM is versatile and meets the practical requirements for various applications.

\begin{table}[]
\center
\resizebox{\linewidth}{!}{
\begin{tabular}{c|cccc}
\toprule
           & \multicolumn{2}{c}{ResNet-18} & \multicolumn{2}{c}{MobileNets} \\ \midrule
methods    & Acc.      & AIS     & Acc.    & AIS   \\ \hline
Full prec. & 88.72     & \textbackslash{}  & 85.81      & \textbackslash{}           \\
LSQ+SGD    & 88.47     & 1720(191\%)    & 83.44   & 1740(201\%)          \\
LSQ+SAM    & 89.46     & 902(100\%)     & 83.83   & 865(100\%)           \\
LSQ+vSAM   & 89.01     & 1487(165\%)    & 83.80   & 1329(154\%)          \\ \bottomrule
\end{tabular}
}
\caption{Results of optimizing LSQ using vSAM on the Cifar-10 dataset.}
\label{table_LSQ}
\end{table}

\section{Conclusions}
In this paper, we discover that the gradient of SAM can be decomposed into the gradient of SGD and the PSF.
To enhance optimization efficiency, we transform the optimization process of SAM into gradient sampling and gradient reuse of the PSF.
We suggest that if the PSF changes slightly, it can be replaced by the previous calculated PSF to reduce the calculation.
This allows most optimization iterations to avoid performing two forward and backward, as in the case of SAM.
We propose an adaptive and efficient optimization method called vSAM based on the variance and gradient norm values of the PSF.
vSAM adaptively adjusts the gradient sampling rate of the PSF to enhance optimization efficiency and reuses the PSF in non-sampling iterations.
Experimental results show that vSAM achieves similar generalization performance as SAM and has a faster optimization speed.

\bibliographystyle{named}
\bibliography{ijcai24}

\end{document}